\documentclass[runningheads]{llncs}
\usepackage[T1]{fontenc}
\usepackage{graphicx}
\usepackage{booktabs}
\usepackage[misc]{ifsym}

\usepackage{times}
\usepackage{amsmath, amssymb}
\usepackage{enumitem}
\usepackage{hyperref}
\usepackage{multirow}

\usepackage[english]{babel}
\addto\extrasenglish{
}%
\addto\extrasngerman{%
}%

\begin{document}
\title{TaskFusion: Continual Anomaly Detection for Heterogeneous Tabular Data}

\author{Dayananda Herurkar\inst{1,2} 
\and
Federico Raue\inst{1} 
\and
Joachim Folz\inst{1}
\and 
Jörn Hees\inst{1,3} 
\and
Andreas Dengel\inst{1,2} 
}

\authorrunning{Herurkar et al.}

\institute{German Research Center for Artificial Intelligence (DFKI), Kaiserslautern, Germany \and
RPTU Kaiserslautern-Landau, Germany \and
Hochschule Bonn-Rhein-Sieg (H-BRS), Germany \\
\email{firstname.lastname@dfki.de}}

\maketitle              
\begin{abstract}
Continual anomaly detection in tabular data is challenging and remains largely underexplored, particularly in settings with heterogeneous feature schemas, distribution shifts, and severe class imbalance. 
In many real-world applications, data arrive sequentially from diverse domains, rendering conventional continual learning methods ineffective due to their reliance on a fixed input space.
We propose a continual learning (CL) method, which can overcome these challenges and continually learn from different tasks.
Our method consists of three main parts: our AGF model,  Taskfusion augmentation, and outlier exposure.
The \textit{AGF-model} maps task-specific features into a shared space, then aligns distributions to reduce representation drift, and learns anomaly decision boundaries in the aligned space. 
To improve stability, we introduce \textit{ Taskfusion augmentation}, combining boundary-aware interpolation within tasks to refine the model anomaly boundaries and cross-task mixing to transfer anomaly structure across datasets. 
To handle class imbalance and memory constraints, we employ tabular dataset distillation to store compact synthetic replay samples, which are jointly used with augmented data in an \textit{outlier exposure} objective for robust anomaly detection.  
We evaluate the approach on 21 heterogeneous datasets across multiple domains. 
Results show that our approach substantially improves continual anomaly detection performance over sequential fine-tuning and other CL baselines while reducing catastrophic forgetting and maintaining stable detection across heterogeneous datasets.

\end{abstract}
\section{Introduction}

Outlier detection in tabular data plays a critical role in many real-world applications, including fraud detection, system monitoring, healthcare analytics, and industrial inspection \cite{Fin-Fraud1}. 
In practical deployments, however, data rarely arrive as a single static dataset. 
Instead, new tabular datasets emerge sequentially from different sources or domains, each characterized by distinct feature schemas, statistical distributions, and anomaly patterns. 
This setting naturally gives rise to 
a situation where a model must incrementally or continuously incorporate new knowledge while preserving previously learned behavior.
A key opportunity in continual anomaly detection lies in aggregating outlier information across datasets. 
Anomalies observed in different domains often capture complementary failure modes, and combining such signals can lead to more robust and generalizable detectors. 
However, realizing this potential is particularly challenging for heterogeneous tabular data. 
As a result, 
joint training is often infeasible, preventing effective knowledge sharing across datasets.
While effective in homogeneous settings, existing continual learning (CL) strategies do not address schema mismatch or cross-domain anomaly representation inconsistency. 
The challenge is further amplified in anomaly detection, where labeled anomalies are scarce, with large class imbalance, and decision boundaries are sparse and unstable. 
Consequently, naive sequential training rapidly forgets earlier anomaly patterns, while standard replay methods struggle with severe class imbalance and storage constraints.

Existing continual anomaly detection approaches typically assume homogeneous feature spaces or operate within a single domain, making them unsuitable for heterogeneous settings where feature schemas and distributions differ across tasks \cite{faber2024lifelong,frikha2021arcade}. 
Moreover, standard replay mechanisms struggle under severe class imbalance and limited memory, limiting their ability to preserve anomaly knowledge over long sequences \cite{rebuffi2017icarl}.
In this work, we address these limitations by shifting the focus from feature-level transfer to representation-level knowledge accumulation. 
We propose a continual learning framework called \textit{AGF}, that first maps heterogeneous tabular inputs into a shared latent space and aligns task-specific distributions, enabling anomaly knowledge to be combined across datasets with different sets of feature spaces. 
Furthermore, we introduce TaskFusion augmentation, which improves robustness through boundary-aware interpolation to refine sparse anomaly decision regions within tasks and cross-task latent mixing to transfer anomaly structure across domains. 
To further stabilize learning under class imbalance and memory constraints, we integrate tabular dataset distillation \cite{Tab-Distillation} as a compact replay mechanism and leverage outlier exposure to reinforce anomaly discrimination. 
Together, these components enable scalable, continual anomaly detection without requiring access to raw historical data.
This work makes the following contributions:
\begin{itemize}
    \item We propose a continual anomaly detection approach called AGF for heterogeneous tabular data.
    \item We introduce TaskFusion augmentation, combining boundary-aware interpolation and cross-task mixing to improve robustness and anomaly generalization.
    \item We integrate tabular dataset distillation to enable efficient replay under limited memory and to overcome class imbalance constraints.
\end{itemize}



\section{Related Work}

\paragraph{Continual Learning:}
Continual learning (CL) studies how to learn from a sequence of tasks without catastrophic forgetting.
A prominent direction constrains parameter drift, such as Elastic Weight Consolidation (EWC), which penalizes updates to parameters estimated to be important for previously learned tasks \cite{kirkpatrick2017ewc}.
A complementary direction performs functional regularization through distillation, e.g., Learning without Forgetting (LwF), which preserves prior behavior by matching outputs of the previous model while training on new-task data \cite{li2016lwf}.
Replay-based CL mitigates forgetting by maintaining a memory of past examples and interleaving them during training.
iCaRL popularized exemplar-based rehearsal for incremental representation learning \cite{rebuffi2017icarl}. 
Furthermore, continual and lifelong anomaly detection have been explored through unlearning-based mechanisms~\cite{du2019unlearning}, sequential one-class formulations such as ARCADe~\cite{frikha2021arcade}, and broader evaluations of lifelong anomaly learning protocols~\cite{faber2024lifelong}.

\paragraph{Anomaly Detection for Tabular Data:}
Recent advances in tabular anomaly detection have shifted toward deep learning and self-supervised representation learning to better capture complex feature interactions. 
Reconstruction-based methods remain widely used, but modern variants enhance them with retrieval or contextual mechanisms, such as retrieval-augmented models that leverage sample dependencies to improve reconstruction quality \cite{retrieval_ad}, \cite{RECol}, \cite{Herurkar-ijcnn23}. 
Self-supervised learning has become a prominent direction, with masked modeling approaches (e.g., masked cell modeling) capturing feature dependencies through reconstruction of partially observed inputs \cite{mcm}, \cite{Herurkar-IEEE}, \cite{Tab-Distillation}.
Transformer-based and context-conditioned models further improve performance by learning feature interactions and adapting to heterogeneous contexts within tabular data \cite{context_ad}. 
Additionally, recent works explore hybrid architectures that combine representation learning with classical principles, such as deep extensions of isolation-based methods \cite{deep_iforest} and influence-based anomaly scoring \cite{tracinad}, \cite{Fin-Fed-OD}. 
Despite these advances, most existing approaches operate in a static setting and assume a fixed feature schema, limiting their applicability to continual and heterogeneous scenarios where datasets evolve over time. 

\paragraph{Outlier Exposure and Data Augmentation:}
Outlier Exposure (OE) enhances anomaly detection by training models against auxiliary outlier datasets, improving robustness to unseen anomalies~\cite{hendrycks2019oe}. 
Such a technique highlights the importance of exposing models to diverse anomaly patterns over time.
Complementary to OE, data augmentation and interpolation strategies have emerged as powerful regularizers. 
MixUp~\cite{zhang2018mixup} and Manifold Mixup~\cite{verma2019manifoldmixup} improve generalization by interpolating samples or hidden representations to smooth decision boundaries. 
For tabular domains, specialized augmentation mechanisms have been proposed, including pseudo-anomaly generation via nearest-neighbor Gaussian Mixup~\cite{dong2024nngmix} and feature corruption–based contrastive learning (SCARF)~\cite{bahri2022scarf}.
While these approaches demonstrate that representation-space corruption and interpolation are effective for tabular learning, they are primarily designed for static or single-dataset settings. Our work integrates outlier exposure with augmentation within a continual learning framework, enabling progressive aggregation of anomaly knowledge across heterogeneous tabular datasets.

\section{Approach}


We consider a CL setting consisting of a sequence of tabular tasks $\mathcal{T}_1, \mathcal{T}_2, \dots, \mathcal{T}_T$, where each task $\mathcal{T}_t = \{X_t, Y_t\}$ corresponds to a dataset with feature dimension $d_t$.
Each sample $x \in \mathbb{R}^{d_t}$ is associated with a binary label $y \in \{0,1\}$ indicating inlier (normal) or outlier (anomaly).
Unlike continual learning in homogeneous settings, feature schemas can differ across tasks ($d_t \neq d_{t'}$).
The objective is to learn a unified anomaly detector that incrementally integrates knowledge from heterogeneous datasets while preventing catastrophic forgetting.
Formally, the model learns a function 
$f_\Theta : \mathbb{R}^{d_t} \rightarrow [0,1]$,
shared across all tasks, which predicts anomaly probability.
The main challenge is therefore twofold:
(1) aligning heterogeneous tabular schemas into a shared representation space, and
(2) maintaining stable anomaly detection under continual distribution shifts.





\subsection{AGF Model}
\begin{figure}[]
\centering
\includegraphics[width=\linewidth]{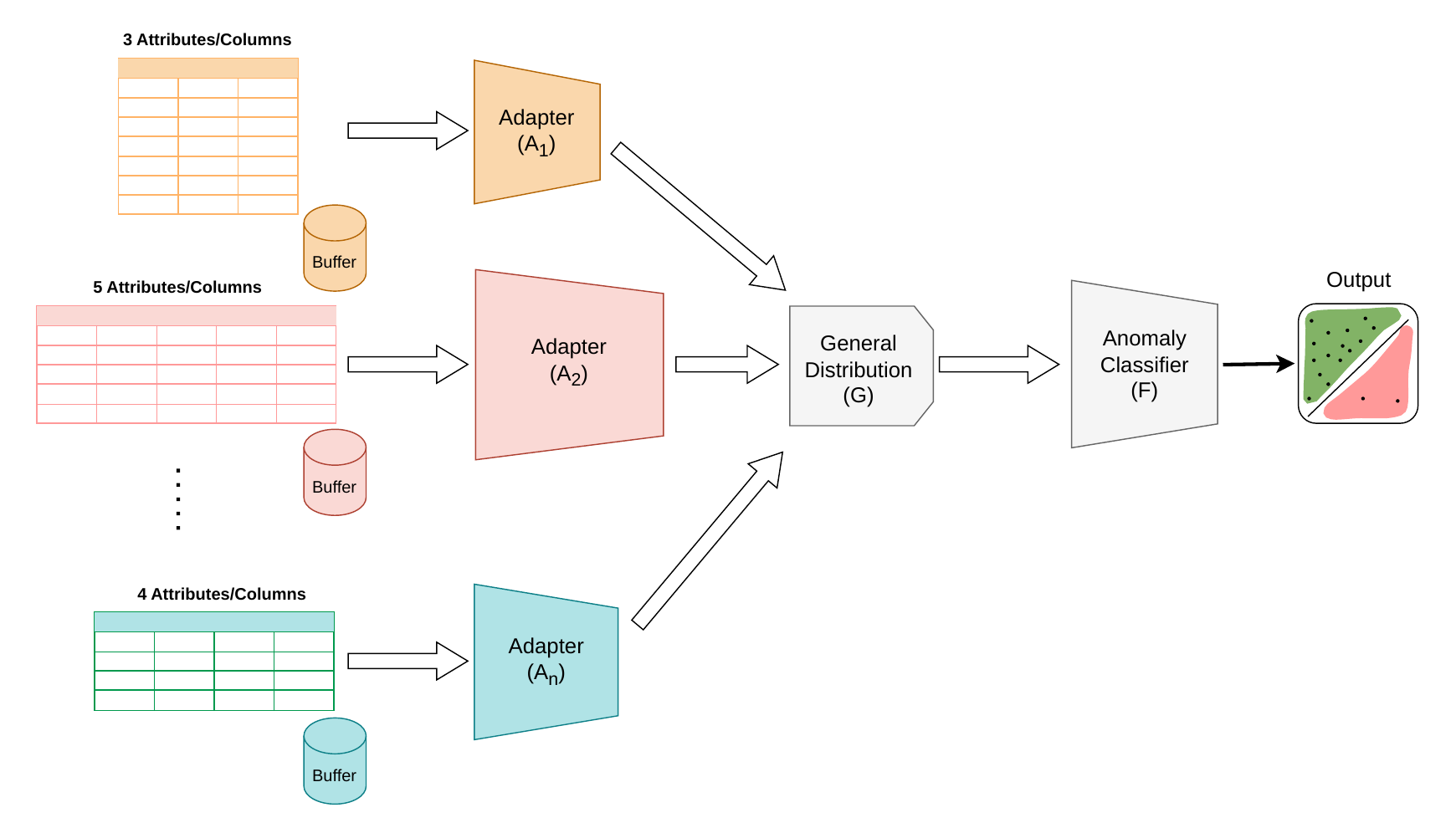}
\caption{AGF architecture for heterogeneous continual anomaly detection.
Task-specific tabular datasets with different feature dimensions are learned using AGF to perform anomaly prediction by knowledge accumulation without shared raw samples. After each task, the buffer is created via tabular dataset distillation to store compact synthetic samples and added to the replay memory that is used in subsequent tasks.}
\label{fig:agf}
\end{figure}

The proposed framework follows a model architecture composed of three stages, referred to as AGF (see \autoref{fig:agf}): 
\begin{equation}
    f_\Theta(X_t) = F_{\theta_f}(G_{\theta_g}(A^{t}_{\theta_a}(X_t)))
\end{equation}
The parameters $\theta_g$ and $\theta_f$ are shared across tasks and updated sequentially while $\theta_a$ is updated only during the respective task.
The task-specific feature spaces are mapped into a shared latent space via a learnable schema adapter.
While raw input dimensionalities differ across tasks, the downstream modules operate in a fixed latent space.

\paragraph{Schema Adapter (A):}
The adapter maps heterogeneous tabular inputs into the same latent dimensionality $d_z$ ie., $A_{\theta_a}: \mathbb{R}^{d_t} \rightarrow \mathbb{R}^{d_z}$. This module enables learning across datasets with different feature dimensions by performing schema-level harmonization.

\paragraph{General Distribution (G):}
The distribution alignment component maps task-specific embeddings into a unified latent distribution $G_{\theta_g}: \mathbb{R}^{d_z} \rightarrow \mathbb{R}^{d_g}$.
The alignment emerges implicitly through joint optimization across tasks, where shared parameters are updated to minimize anomaly classification loss across heterogeneous domains.
By calibrating representations across tasks, the model reduces domain discrepancies and enables knowledge transfer between datasets.
Given an input $x$, its latent embedding becomes $g = G_{\theta_g}(A_{\theta_a}(x))$.

\paragraph{Anomaly Classifier (F):}
The final module performs binary anomaly classification based on aligned representations.
It learns a robust decision boundary separating normal and anomalous samples while adapting continuously as new tasks arrive.
$F_{\theta_f}: \mathbb{R}^{d_g} \rightarrow \mathbb{R}^{2}$ produces logits used for binary anomaly classification.
The anomaly probability is computed via softmax.

Together, these components allow the model to accumulate knowledge across heterogeneous domains without requiring identical feature schemas.
The task loss is defined as standard cross-entropy:
$
L_{\text{task}} = \frac{1}{N} \sum_{i=1}^{N} \mathrm{CE}\big(f_{\Theta}(x_i),\, y_i\big).
$
The architecture enables combining outlier information from multiple datasets through a shared latent representation, which would otherwise be infeasible in raw feature space. 

\subsection{Augmentation Techniques}

To improve robustness and mitigate catastrophic forgetting, we introduce TaskFusion Augmentation, which operates in the learned latent space (G).






\paragraph{Within-Task Boundary-Aware Augmentation:}
Samples with uncertain predictions, typically located near the anomaly decision boundary, are identified within the current task.
Interpolation among these samples generates additional training points that densify boundary regions.
This encourages smoother decision surfaces and improves calibration in sparse anomaly regimes.
Within each task, uncertain samples located near the anomaly boundary are identified using prediction confidence: $\mathcal{U}_t = \{x_i \mid \tau_1 < p_{\text{anom}}(x_i) < \tau_2 \}$ where $\tau_1$ and $\tau_2$ are adaptive quantile-based thresholds over predicted scores. 
Pairs $(\alpha_i, \alpha_j)$ drawn from $\mathcal{U}_t$ are then combined using a MixUp with $\lambda \in (0, 1)$: \mbox{$\tilde{aug}_{ij} = \lambda \alpha_i + (1-\lambda) \alpha_j$}.
Mixing is restricted to samples with identical labels (anomalies with anomalies or normal with normal samples) to preserve semantic consistency.
This augmentation densifies sparse boundary regions and encourages smooth decision surfaces.


\paragraph{Cross-Task Augmentation:}
To transfer knowledge between tasks, embeddings from the current and previous tasks are combined in latent space.
Nearest-neighbor similarity is used to identify compatible samples, and convex interpolation produces hybrid representations.
Pre- and post-consistency checks ensure that generated samples remain semantically valid and consistent with anomaly labels.
To combine anomaly information across datasets, embeddings from the current task and previous tasks are mixed.
For a current embedding $\alpha_i^{(t)}$, a compatible historical embedding is selected via nearest-neighbor search: $j = \arg\min_k
\| \alpha_i^{(t)} - \alpha_k^{(t-1)} \|_2$.
MixUp then generates $\tilde{aug}_{ij} = \lambda \alpha_i^{(t)} + (1-\lambda) \alpha_j^{(t-1)}$.
Furthermore, compatibility constraints ensure safe mixing with $\text{cos}(\alpha_i^{(t)}, \alpha_j^{(t-1)}) \ge \tau_{\text{cos}}, \quad \|\alpha_i^{(t)}-\alpha_j^{(t-1)}\|_2 \le \epsilon$.
where, the cosine similarity enforces directional alignment, while Euclidean proximity controls local density compatibility.
This mechanism enables aggregation of outlier characteristics across heterogeneous datasets, which is otherwise infeasible in raw tabular feature space.



\subsection{Outlier Exposure Loss}
Augmented samples are incorporated as additional training signals through an outlier exposure mechanism.
Synthetic samples generated via augmentation, together with distilled samples of replay memory from previous tasks, provide diverse anomaly-related supervision during training.
This strategy allows the model to reuse historical anomaly information without requiring raw data storage.
As learning progresses, the detector benefits from accumulated exposure to heterogeneous anomaly patterns originating from multiple datasets.
Augmented and replay synthetic samples are incorporated using a supervised outlier exposure (OE) objective.
Let $z_i = F_{\theta_f}(aug_i)$ denote classifier logits and $y_i \in \{0,1\}$ the anomaly label.
The weighted cross-entropy loss is defined as $\ell_i = \text{CE}(z_i, y_i)$.
To emphasize anomaly samples, anomaly instances ($y_i$=1) are assigned a weight($w_i$) 0.9, while normal instances ($y_i$=0) receive unit weight.
The OE loss becomes $\mathcal{L}_{\text{OE}} = \frac{1}{N} \sum_{i=1}^{N} w_i \, \ell_i$.
This formulation corresponds to supervised OE using weighted binary classification, where anomaly samples receive stronger optimization pressure.
The final optimization objective combines task learning, augmentation, and outlier exposure:


\begin{equation}
    \mathcal{L}_{\text{total}}
    =
    \mathcal{L}_{\text{task}}
    +
    \lambda_{\text{oe}}\mathcal{L}_{\text{OE}}.
\end{equation}

This objective promotes stable continual learning by aligning heterogeneous representations, refining decision boundaries, and aggregating anomaly information across tasks.

\section{Experimental Setup}

\subsection{Datasets}

We evaluate the proposed framework on 21 heterogeneous tabular benchmark datasets spanning multiple domains, including finance, image-derived representations, astronautics and sensor monitoring, and document-based tabular features.
The datasets vary substantially in feature dimensionality, sample size, and anomaly ratio. 
Feature dimensions range from low-dimensional structured attributes to high-dimensional derived representations, and anomaly rates vary across domains, creating realistic class imbalance scenarios.
Crucially, feature schemas differ across datasets ($d_t \neq d_{t'}$), making joint training in raw feature space infeasible.
This heterogeneous setting reflects practical deployment conditions where new tabular sources emerge over time with incompatible schemas.
Detailed statistics for all datasets are provided in the Appendix. 

\subsection{Continual Learning Protocol}
Datasets are presented sequentially as tasks $\mathcal{T}_1, \dots, \mathcal{T}_T$.
At each step, the model has access only to current task data and a fixed-size replay memory constructed via tabular dataset distillation \cite{Tab-Distillation}.
After completing task $\mathcal{T}_t$, a distilled subset of size $K$ (instances per class, IPC) is generated with the distribution matching \cite{dm} approach and stored in memory.
Replay capacity remains fixed throughout training.
Unless otherwise specified, results are averaged over multiple random task orderings to reduce ordering bias.
After completing training on each task, the model is evaluated on all observed tasks to measure knowledge retention and forward transfer.
This protocol measures adaptation to new heterogeneous datasets, knowledge retention, and cross-task anomaly accumulation.

\subsection{Baseline Methods}
We compare the proposed framework against several representative baselines:

\noindent \textbf{Independent Training (w/o CL):} Standard anomaly detection models are trained independently on each dataset without any continual learning. The final performance is reported as the average across all datasets/tasks. These methods represent conventional single-domain approaches. \\
\noindent \textbf{Finetune:} Sequential training or finetuning of the model on each task. \\
\noindent \textbf{Multitask:} A non-continual upper-bound setting where all datasets are jointly trained. This provides a reference performance assuming simultaneous access to all data. \\
\noindent \textbf{CaSSLe \cite{cassle}:} A continual self-supervised learning baseline where representation drift is mitigated via a predictor-based distillation mechanism that aligns current representations with their past states. \\
\noindent \textbf{EDSR \cite{edsr}:} We adopt Effective Data Selection and Replay (EDSR), a CL method that selects representative samples based on entropy and replays them with noise-enhanced knowledge distillation.

To analyze the contribution of individual components, we consider the following ablations of the proposed method: \\
\noindent \textbf{AGF (No Augmentation):} The proposed AGF architecture trained without TaskFusion augmentation. \\
\noindent \textbf{AGF (No Outlier Exposure):} A variant where the outlier exposure is disabled, isolating the impact of augmented and other task synthetic outliers across heterogeneous datasets.
    
All methods use identical architecture, capacity, and optimization settings for fair comparison. The complete implemetation details are described in Appendix. 
Performance is evaluated using Balanced Accuracy, ROC-AUC, and PR-AUC, along with task-wise performance evolution to assess how well models retain past knowledge while adapting to new tasks. We report average continual performance and forgetting, where high final performance and low forgetting indicate effective knowledge retention and accumulation across heterogeneous tasks. Details on evaluation metrics can also be found in Appendix. 

\section{Results}

We evaluate our approach under two research questions:


\noindent \textbf{Research Question 1:} Can AGF effectively combine outlier information across heterogeneous tabular datasets for continual learning? \\
\noindent \textbf{Research Question 2:} Do TaskFusion and outlier exposure (OE) stabilize learning under tasks with long sequences and mitigate catastrophic forgetting?

\subsection{Overall Performance on Heterogeneous Benchmarks}

\autoref{tab:main_results} summarizes the performance of continuous anomaly detection across 21 heterogeneous tabular datasets grouped by domain, with results averaged over three independent runs. 
Sequential finetuning exhibits severe degradation across all domains,
confirming that naive sequential learning fails under schema mismatch and strong distribution shifts. 
Multitask training provides a strong upper bound but is impractical; notably, the proposed AGF framework with TaskFusion and outlier exposure (OE) substantially narrows this gap and even matches or exceeds multitask performance in some cases. 
Representation alignment via AGF already improves stability over finetuning, and replacing random replay with dataset distillation further enhances performance. 
Incorporating OE yields consistent gains across all metrics (e.g., Balanced Accuracy improves),  
highlighting the importance of exposing the model to diverse anomaly structures. 
TaskFusion augmentation provides additional improvements, with combined within-task and cross-task augmentation achieving the best results across nearly all settings, particularly in Image and Finance domains, where feature heterogeneity is high. 
Gains are more moderate in Astronautics and Document domains, where baseline performance is already strong or task diversity is limited. 
Overall, the results demonstrate that integrating representation alignment, augmentation, and OE enables effective knowledge accumulation and robust continual anomaly detection across heterogeneous tabular datasets.


\begin{table}[t]
\centering
\caption{Continual anomaly detection performance across 21 heterogeneous tabular datasets.
Best continual results are shown in bold. RS: Replay with Random Sampling, DD: Replay with Dataset Distillation, WT: WithinTask Augmentation, CT: CrossTask Augmentation.}
\label{tab:main_results}
\resizebox{\textwidth}{!}{
\begin{tabular}{@{}lrccccccccccc@{}}
\toprule
\textbf{Field} & \multicolumn{1}{c}{\textbf{Tasks}} & \textbf{w/o CL} & \textbf{Finetune} & \textbf{Multitask} & \multicolumn{1}{l}{\textbf{CaSSLe}} & \multicolumn{1}{l}{\textbf{EDSR}} & \textbf{\begin{tabular}[c]{@{}c@{}}AGF+\\ RS\end{tabular}} & \textbf{\begin{tabular}[c]{@{}c@{}}AGF+\\ DD\end{tabular}} & \textbf{\begin{tabular}[c]{@{}c@{}}AGF+OE \\ +DD\end{tabular}} & \textbf{\begin{tabular}[c]{@{}c@{}}AGF+OE\\ +WT \\ +DD\end{tabular}} & \textbf{\begin{tabular}[c]{@{}c@{}}AGF+OE\\ +CT \\ +DD\end{tabular}} & \textbf{\begin{tabular}[c]{@{}c@{}}AGF+OE\\ +WT+CT\\ +DD\end{tabular}} \\ \midrule
\multicolumn{13}{c}{\textbf{Balanced Accuracy}} \\ \midrule
\textbf{All} & 21 & 78.74 & 51.12 & 79.73 & 63.52 & 67.43 & 72.97 & 75.09 & 80.37 & 82.99 & 83.34 & \textbf{83.92} \\
\textbf{Image} & 6 & 71.55 & 53.02 & 72.80 & 54.27 & 62.75 & 69.54 & 72.06 & 74.72 & 75.87 & 75.29 & \textbf{76.91} \\
\textbf{Finance} & 5 & 70.92 & 51.92 & 71.01 & 56.57 & 58.43 & 67.82 & 69.09 & 72.93 & 72.68 & 73.10 & \textbf{74.75} \\
\textbf{Astronautics} & 3 & 92.12 & 66.21 & \textbf{94.37} & 68.43 & 75.83 & 88.56 & 92.55 & 92.91 & 92.54 & 92.87 & 93.77 \\
\textbf{Document} & 2 & 81.02 & 56.65 & \textbf{83.05} & 62.70 & 64.70 & 73.32 & 76.20 & 76.99 & 78.64 & 78.39 & 78.43 \\ \midrule
\multicolumn{13}{c}{\textbf{PR-AUC}} \\ \midrule
\textbf{All} & 21 & 68.14 & 32.88 & \textbf{70.09} & 46.27 & 48.91 & 56.97 & 62.76 & 65.26 & 66.66 & 66.76 & 68.85 \\
\textbf{Image} & 6 & 54.74 & 32.96 & \textbf{55.49} & 27.01 & 40.26 & 46.48 & 50.32 & 51.14 & 52.44 & 52.12 & 52.83 \\
\textbf{Finance} & 5 & 54.56 & 35.84 & \textbf{55.85} & 35.91 & 40.71 & 46.14 & 51.98 & 52.60 & 53.85 & 53.85 & 55.10 \\
\textbf{Astronautics} & 3 & 95.61 & 66.53 & \textbf{96.56} & 47.31 & 41.23 & 85.84 & 94.87 & 94.78 & 95.30 & 95.89 & 96.36 \\
\textbf{Document} & 2 & 76.15 & 43.69 & \textbf{77.67} & 46.72 & 56.55 & 62.28 & 67.04 & 69.11 & 69.03 & 70.42 & 70.88 \\ \midrule
\multicolumn{13}{c}{\textbf{ROC-AUC}} \\ \midrule
\textbf{All} & 21 & 86.26 & 51.50 & 88.88 & 69.58 & 71.46 & 80.52 & 87.00 & 88.28 & 89.53 & 89.53 & \textbf{90.54} \\
\textbf{Image} & 6 & 84.00 & 59.43 & 85.06 & 65.94 & 74.91 & 80.86 & 84.60 & 84.72 & 84.65 & 84.13 & \textbf{85.20} \\
\textbf{Finance} & 5 & 81.78 & 53.47 & 82.30 & 68.69 & 72.83 & 74.47 & 81.35 & 82.14 & 81.34 & 82.02 & \textbf{82.41} \\
\textbf{Astronautics} & 3 & 97.43 & 71.19 & 98.63 & 62.85 & 60.46 & 90.25 & 97.95 & 98.02 & 98.66 & 98.24 & \textbf{98.85} \\
\textbf{Document} & 2 & 88.94 & 61.74 & \textbf{90.95} & 65.46 & 75.36 & 79.34 & 85.80 & 85.88 & 87.31 & 87.37 & 90.88 \\ \bottomrule
\end{tabular}
}
\end{table}

\subsection{Ablation Study}






\noindent \textbf{Impact of Augmentation Components:}
We further analyze the contribution of individual components by progressively enabling augmentation mechanisms within the AGF framework (in \autoref{tab:main_results}). While representation alignment with tabular dataset distillation already improves over finetuning, it remains insufficient in long sequences (e.g., 
in the 21-task setting), indicating accumulated boundary drift. Introducing within-task boundary-aware augmentation significantly improves performance 
by densifying uncertain regions and stabilizing anomaly decision boundaries. Cross-task augmentation further enhances results, 
with notable gains in PR-AUC, suggesting improved detection of rare anomalies through shared representation exposure across tasks. The combination of both mechanisms achieves the best and most consistent performance across domains, demonstrating complementary effects: within-task augmentation refines local decision boundaries, while cross-task mixing facilitates knowledge transfer across heterogeneous datasets. Overall, these results confirm that augmentation acts as an effective structural regularizer, mitigating representation drift and improving robustness in long continual learning sequences.

\noindent \textbf{Boundary Selection Analysis:}
We analyze the influence of the boundary-quality (BQ) threshold used for selecting samples in task-wise augmentation. 
\autoref{tab:bq} reports the resulting Balanced Accuracy, True Positive Rate (TPR), and True Negative Rate (TNR).
The results reveal a clear trade-off between anomaly sensitivity and normal-class preservation. 
Lower BQ thresholds favor conservative augmentation, yielding very high TNR values (e.g., $95.27$ at BQ $=0.6$) but comparatively lower TPR, indicating limited improvement in detecting difficult anomalies. 
As the threshold increases, augmentation focuses more strongly on uncertain boundary samples, leading to substantial gains in TPR, which steadily improves and reaches its maximum at BQ $=0.95$.
Balanced Accuracy peaks at a moderate threshold (BQ $=0.8$), suggesting that selecting samples near but not exclusively at the decision boundary provides the best overall balance between detecting anomalies and maintaining stable inlier classification. 
Beyond this point, further increasing the threshold continues to improve anomaly sensitivity but causes a noticeable reduction in TNR, indicating overemphasis on ambiguous regions.
Overall, these results confirm that boundary-aware augmentation is most effective when applied to moderately uncertain samples. 
Rather than uniformly augmenting the dataset, focusing on informative boundary regions enables improved anomaly detection while preserving overall classification stability.

\begin{table}[hbt]
\centering
\caption{Effect of boundary-quality (BQ) threshold in task-wise augmentation. 
Moderate thresholds provide the best balance between anomaly detection (TPR) and inlier preservation (TNR), yielding the highest Balanced Accuracy at $\text{BQ}=0.8$.}
\label{tab:bq}
{
\setlength{\tabcolsep}{6pt}
\begin{tabular}{lccccc}
\toprule
Metric (BQ) & 0.6 & 0.7 & 0.8 & 0.9 & 0.95 \\
\midrule
Bal Acc & 75.68 & 77.95 & \textbf{78.92} & 77.18 & 78.25 \\
TPR  & 56.09 & 62.95 & 67.94 & 68.44 & \textbf{74.21} \\
TNR & \textbf{95.27} & 92.95 & 89.91 & 85.92 & 82.31 \\
\bottomrule
\end{tabular}
}
\end{table}

\subsection{Influence of Replay Method and Capacity}

\begin{figure}[]
\centering
\includegraphics[width=\linewidth]{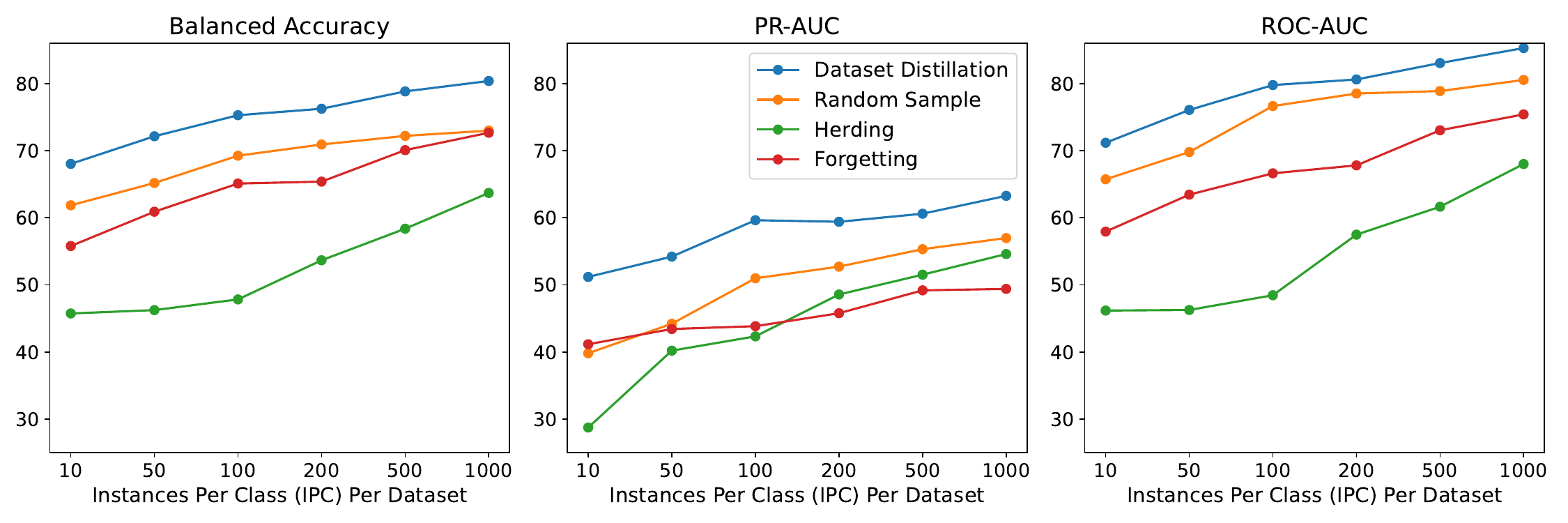}
\caption{Influence of replay capacity measured by instances per class (IPC). 
Increasing IPC consistently improves performance across Balanced Accuracy, PR-AUC, and ROC-AUC. 
Dataset distillation achieves the strongest results across all capacities, while herding is highly sensitive to memory size and random sampling shows moderate gains.}
\label{fig:replay_memory}
\end{figure}

\autoref{fig:replay_memory} evaluates replay capacity by varying instances per class (IPC) and under different replay strategies.
Across all metrics, increasing IPC consistently improves performance.
Dataset distillation achieves the strongest results at every capacity level, indicating that replay quality is as important as replay quantity \cite{Tab-Distillation}.
Random sampling improves steadily with IPC, but remains below distillation.
Herding is highly sensitive to small memory budgets, performing poorly at low IPC and improving only when sufficient capacity is available.
In contrast, dataset distillation maintains strong performance even under moderate memory constraints.
These findings demonstrate that structured data representations in replay memory enable efficient knowledge retention and improved anomaly discrimination.

\section{Conclusion}

We addressed continual anomaly detection on heterogeneous tabular data, where datasets differ in feature dimensionality, distribution, and anomaly structure. 
Unlike continual learning in homogeneous settings, raw feature spaces are incompatible, making direct data reuse or joint optimization infeasible.
We introduced a continual learning framework that enables progressive aggregation of outlier knowledge across heterogeneous domains. 
The approach integrates three complementary components: 
(i) the AGF architecture for schema adaptation and distribution alignment in a shared latent space, 
(ii) TaskFusion augmentation combining boundary-aware within-task interpolation and cross-task latent mixing, and 
(iii) outlier exposure with distilled replay representations and augmented samples to refine anomaly boundaries without storing raw historical data.
Experiments on 21 benchmark datasets demonstrate improved stability and detection performance over sequential fine-tuning and non-continual baselines. 
The framework achieves reduced forgetting, knowledge accumulation across domains, and robust behavior under long heterogeneous task sequences.
More broadly, our results suggest that heterogeneous continual anomaly detection is best formulated as progressive outlier exposure in representation space rather than parameter-constrained preservation in raw feature space. 
By combining alignment, augmentation, and structured replay, our approach provides a practical and scalable solution for continual anomaly detection across diverse structured data sources.

While the proposed framework improves continual anomaly detection on heterogeneous tabular data, challenges remain in handling extremely long task sequences. 
Additionally, the current approach relies on implicit feature alignment and fixed augmentation strategies, which may limit robustness under diverse or highly dissimilar domains.
Future work will focus on adaptive augmentation, semantic-aware alignment, and extending the framework to fully online or streaming settings for scalable real-world deployment.

\subsubsection{\ackname} This work was supported by the BMBF project Albatross(Grant 01IW24002). 

%
%
%
\bibliographystyle{splncs04}
\bibliography{references}
%

\appendix
\section{Appendix:}

\begin{figure}[]
\centering
\includegraphics[width=1\linewidth]{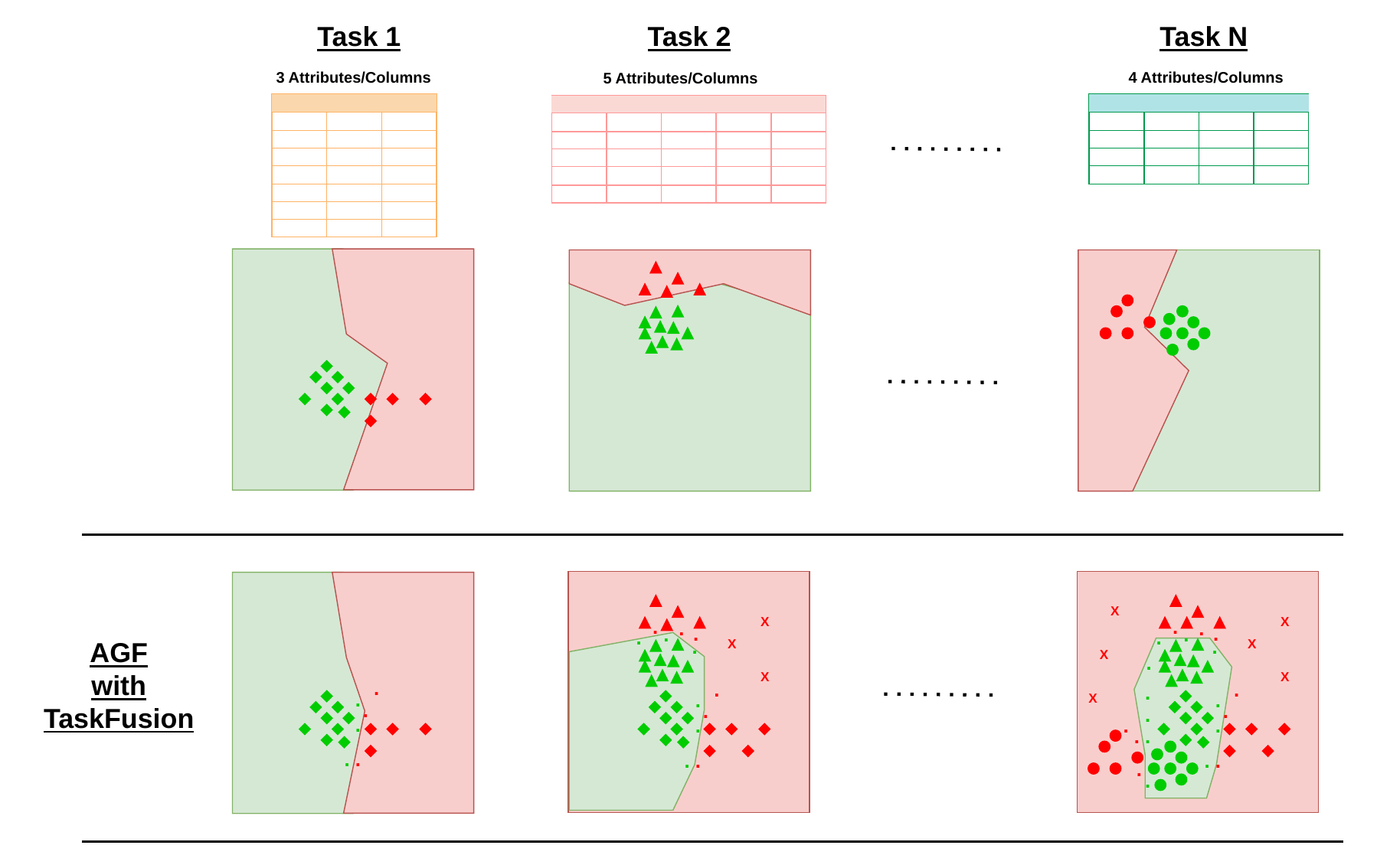}
\caption{Conceptual illustration of continual anomaly detection across heterogeneous tabular datasets. Each task corresponds to a dataset with a different feature schema and decision boundary. In standard sequential learning (top), models trained independently on each task cannot effectively transfer anomaly knowledge across datasets. The proposed AGF with TaskFusion framework (bottom) aligns heterogeneous feature spaces into a shared representation and augments training through boundary-aware interpolation (represented as .) and cross-task mixing (represented as x). This enables the model to accumulate anomaly knowledge across tasks, improving decision boundary stability and detection performance under heterogeneous continual learning. Here, green represents the normal (inlier) space and red represents the anomaly space.}
\label{fig:agf_taskfusion}
\end{figure}

To further understand the behavior of the proposed framework under long continual learning sequences, we provide additional qualitative and distributional analyses of the learned latent space and anomaly scoring function. The \autoref{fig:agf_taskfusion} describes the idea of our approach.

\subsection{Experimental Datasets}  \label{subsec:datasets}
For our study, we utilize a comprehensive selection of datasets shown in \autoref{tab:datasets}, primarily sourced from ADBench \cite{han2022adbench}, one of the most extensive benchmarks for tabular outlier detection.

\begin{table}[tbh]
\centering
\caption{Description of the \href{}{21} tabular outlier detection datasets used in our study, in terms of Sample counts, Columns, Outlier counts/\% and Fields.}
\begin{tabular}{@{}rlrrrrl@{}}
\toprule
 & \multicolumn{1}{l}{\textbf{Dataset}} & \multicolumn{1}{r}{\textbf{Samples}} & \multicolumn{1}{r}{\textbf{Cols}} & \multicolumn{1}{r}{\textbf{Outliers}} & \multicolumn{1}{r}{\textbf{Outlier\%}} & \multicolumn{1}{l}{\textbf{Field}} \\ \midrule
1 & ALOI & 49534 & 27 & 1508 & 3.04 & Image \\
2 & backdoor & 95329 & 196 & 2329 & 2.44 & Network \\
3 & campaign & 41188 & 62 & 4640 & 11.26 & Finance \\
4 & celeba & 202599 & 39 & 4547 & 2.24 & Image \\
5 & census & 299285 & 500 & 18568 & 6.21 & Sociology \\
6 & cover & 286048 & 10 & 2747 & 0.96 & Botany \\
7 & donors & 619326 & 10 & 36710 & 5.92 & Sociology \\
8 & http & 567498 & 3 & 2211 & 0.38 & Web \\
9 & landsat & 6435 & 36 & 1333 & 20.71 & Astronautics \\
10 & magicgamma & 19020 & 10 & 6688 & 35.16 & Physical \\
11 & satellite & 6435 & 36 & 2036 & 31.63 & Astronautics \\
12 & shuttle & 49097 & 9 & 3511 & 7.15 & Astronautics \\
13 & skin & 245057 & 3 & 50859 & 20.75 & Image \\
14 & SpamBase & 4207 & 57 & 1679 & 39.91 & Document \\
15 & creditdef. & 30000 & 146 & 6636 & 22.12 & Finance \\
16 & adultdata & 48842 & 108 & 11687 & 23.92 & Finance \\
17 & blastchar & 7043 & 20 & 1869 & 26.54 & Image \\
18 & bankmarketing & 45211 & 51 & 5289 & 11.69 & Finance \\
19 & Income & 32561 & 108 & 7841 & 24.08 & Finance \\
20 & shoppers & 12330 & 75 & 1908 & 15.47 & Document \\
21 & shrutime & 10000 & 15 & 2037 & 20.37 & Image \\ \bottomrule
\end{tabular}
\label{tab:datasets}
\end{table}

\subsection{Implementation Details}  \label{subsec: imple_details}
All models are trained using the same optimization settings to ensure fair comparison.
The AGF framework is trained using cross-entropy classification loss combined with outlier exposure objectives.
Models are implemented in PyTorch 2.1 and optimized using Adam optimizer.
Each task is trained for a fixed number of epochs with a consistent batch size and learning rate across methods.
The AGF model is instantiated as a multi-layer perceptron (MLP) architecture for all three components (adapter $A$, shared representation $G$, and classifier $F$), with LeakyReLU activation functions. 
For most datasets, the adapter, representation, and classifier consist of hidden layers of sizes $[256,128]$, $[64,64]$, and $[32]$, respectively. For higher-dimensional domains such as image-derived features, larger configurations $[512,256]$, $[128,128]$, and $[64]$ are used, while for certain datasets (e.g., astronautics and document), deeper classifier heads (e.g., $[32,16,8,4]$) are employed. 
All methods are trained for a fixed number of epochs (1000) with consistent batch size and learning rate across methods to ensure fair comparison. Replay buffers are constructed using dataset distillation with a fixed size of 1000 instances per class (IPC). The replay regularization coefficient is set to $\lambda = 0.95$. 
The AGF framework is trained using cross-entropy classification loss combined with outlier exposure objectives. Latent dimensions $d_z$ and $d_g$, augmentation mixing coefficient $\alpha$, and OE weighting parameter $\omega_{\text{oe}}$ are kept fixed across datasets.
The code and models used in this study will be publicly released upon the publication of this paper to facilitate reproducibility and further research.

\subsection{Evaluation Measure}  \label{subsec:eval_measure} 
Performance is evaluated using Balanced Accuracy, ROC-AUC, and PR-AUC, which are standard for imbalanced settings such as anomaly detection.
Beyond final performance, continual learning methods must balance two competing objectives: retaining knowledge from previously learned tasks while effectively integrating information from new datasets.
To analyze this behavior, we evaluate models using task-wise performance evolution across the learning sequence.
Let $R_{i,j}$ denote the performance (e.g., ROC-AUC or Balanced Accuracy) measured on task $\mathcal{T}_i$ after completing training on task $\mathcal{T}_j$, where $j \ge i$.
The matrix $R \in \mathbb{R}^{T \times T}$ therefore captures how performance evolves as new tasks are learned.

\paragraph{Average Continual Performance:}
The overall continual learning performance after the final task is defined as
\[
\text{Score} = \frac{1}{T} \sum_{i=1}^{T} R_{i,T},
\]
which measures the model’s ability to maintain performance across all observed datasets.

\paragraph{Forgetting Measure:}
Catastrophic forgetting is quantified as the average performance degradation on previously learned tasks:
\[
\text{Forgetting}
=
\frac{1}{T-1}
\sum_{i=1}^{T-1}
\max_{j \in \{i,\dots,T-1\}} R_{i,j}
-
R_{i,T}.
\]
This metric captures how much performance is lost after subsequent tasks are learned.
Low forgetting and high average continual performance indicate stable knowledge retention and successful accumulation of anomaly information across heterogeneous tasks.

\subsection{Forgetting and Knowledge Accumulation}

\begin{figure}[]
\centering
\includegraphics[width=\linewidth]{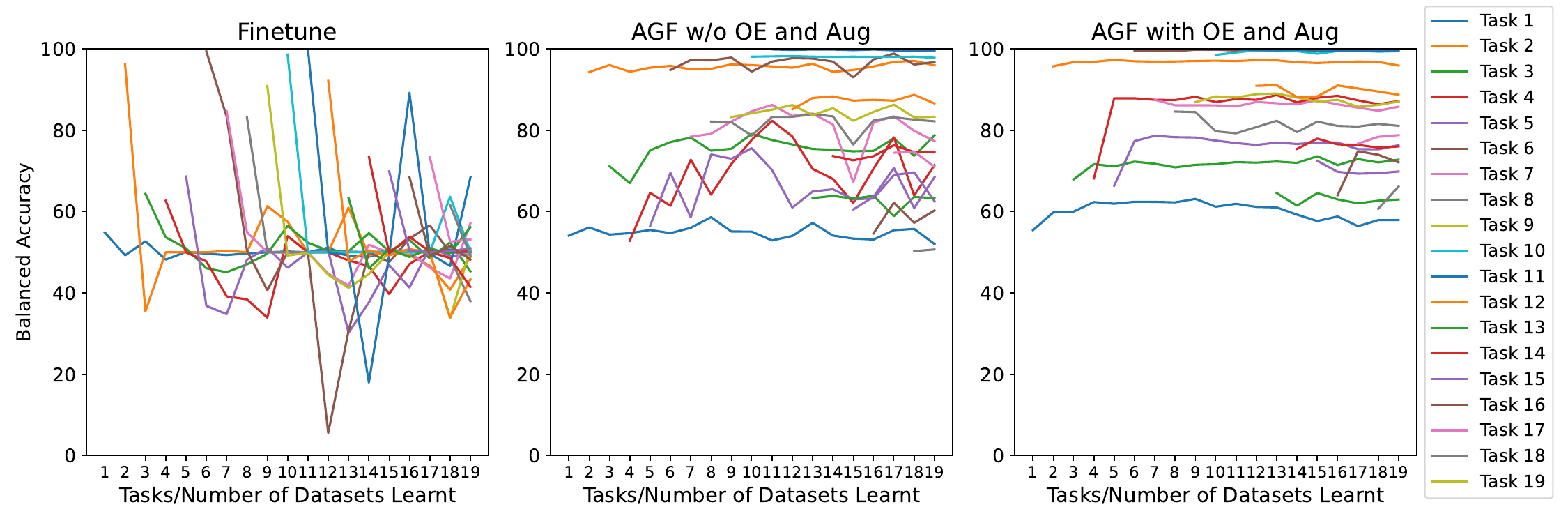}
\caption{Taskwise performance evolution during continual learning. Each curve tracks a task as new datasets are introduced. Finetuning shows repeated drops due to catastrophic forgetting. AGF without OE partially reduces degradation, while AGF with OE and augmentation maintains nearly stable performance across tasks.}
\label{fig:forgetting}
\end{figure}

We further analyze forgetting behavior using taskwise performance evolution and quantitative forgetting measurements. \autoref{fig:forgetting} illustrates taskwise accuracy for all tasks as new datasets are introduced, while \autoref{tab:forgetting} reports average forgetting across domains.
Sequential finetuning shows repeated performance drops whenever a new task is introduced.
This confirms severe interference under heterogeneous schema changes.
AGF without OE and TaskFusion reduces but does not eliminate degradation.
Some tasks still experience noticeable drops as training progresses.
With OE and augmentation enabled, taskwise curves remain nearly flat.
Once a task is learned, its performance remains stable despite subsequent updates.
This indicates strong robustness against catastrophic forgetting.
In \autoref{tab:forgetting}, average forgetting remains close to zero across domains.
Notably, image and astronautics domains exhibit negative forgetting values, indicating slight performance improvement after learning additional tasks.
This suggests knowledge accumulation rather than interference.
Together, these results demonstrate that AGF with TaskFusion enables continual refinement of anomaly representations without strong forgetting.

\begin{table}[t]
\centering
\caption{Average forgetting across domains (lower is better). Values near zero indicate stable retention, while negative values suggest knowledge accumulation. 
}
\label{tab:forgetting}
\begin{tabular}{@{}clccccc@{}}
\toprule
\multirow{3}{*}{\textbf{Forgetting}} & \textbf{Metric} & All & Image & Finance & Astronautics & Document \\ \cmidrule(l){2-7} 
 & \textbf{Balanced Accuracy} & 1.28 & -0.97 & 0.56 & -0.69 & -0.30 \\
 & \textbf{TPR} & 0.18 & -4.45 & -0.45 & -1.37 & -1.78 \\ \bottomrule
\end{tabular}
\end{table}

\subsection{Effect of Long Sequences}

\begin{figure}[]
\centering
\includegraphics[width=\linewidth]{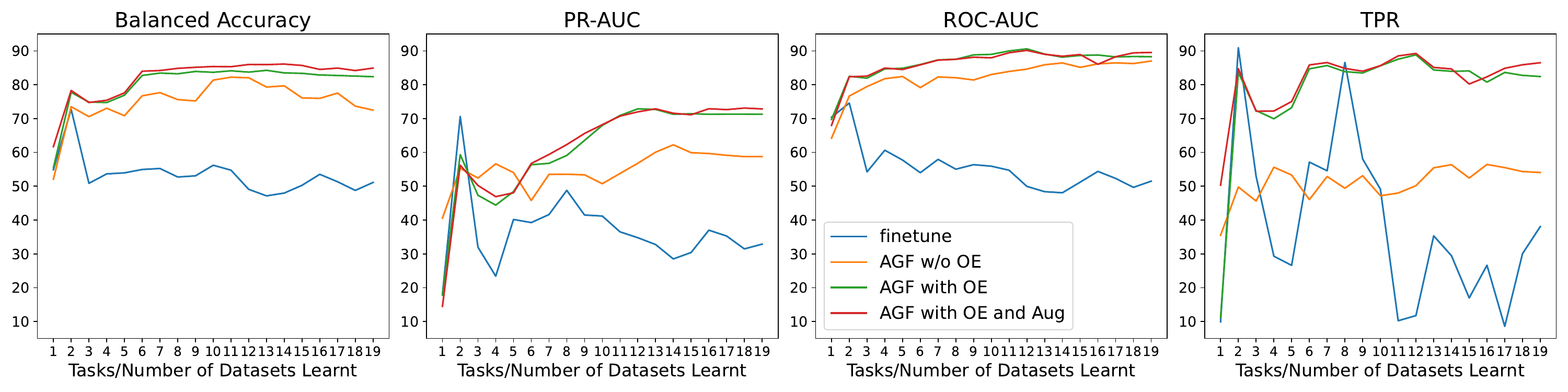}
\caption{Performance evolution over long task sequences. The plots show average performance across previously learned tasks as new datasets are introduced. Finetuning rapidly deteriorates due to catastrophic forgetting. AGF without OE degrades after several tasks, while OE and augmentation stabilize learning and maintain consistently high performance across all metrics.}
\label{fig:long_sequence}
\end{figure}

\autoref{fig:long_sequence} shows the evolution of average performance over all previously learned tasks as the number of tasks or sequential datasets increases. This analysis evaluates the robustness of different approaches under prolonged continual learning for heterogeneous data.
Sequential finetuning rapidly deteriorates as new tasks are introduced. Across all metrics, performance becomes highly unstable after only a few tasks, confirming that naive optimization leads to severe forgetting when feature spaces and data distributions change.
The AGF framework without OE or augmentation performs well during early stages but begins to degrade after approximately 8--9 tasks. This drop is consistently observed across Balanced Accuracy, PR-AUC, ROC-AUC, and TPR, indicating that representation alignment alone is insufficient to prevent accumulated drift in long heterogeneous sequences.
Introducing OE substantially improves stability. AGF with OE maintains nearly constant performance as additional datasets are learned, demonstrating that exposure to broader anomaly structure helps preserve previously acquired decision boundaries.
The most robust behavior is achieved when OE is combined with augmentation. AGF with OE and augmentation shows consistently high and stable performance across the entire sequence, with no noticeable degradation even after learning many datasets. The improvement is particularly evident in all metrics, suggesting better preservation of rare anomaly detection capability over time.
Overall, these results indicate that long-horizon continual learning on heterogeneous tabular data requires both distribution-aware alignment and augmentation-based regularization. While alignment enables knowledge accumulation, augmentation plays a critical role in preventing performance collapse as task length increases.

\subsection{Evolution of the Latent Representation}

\begin{figure}[]
\centering
\includegraphics[width=\linewidth]{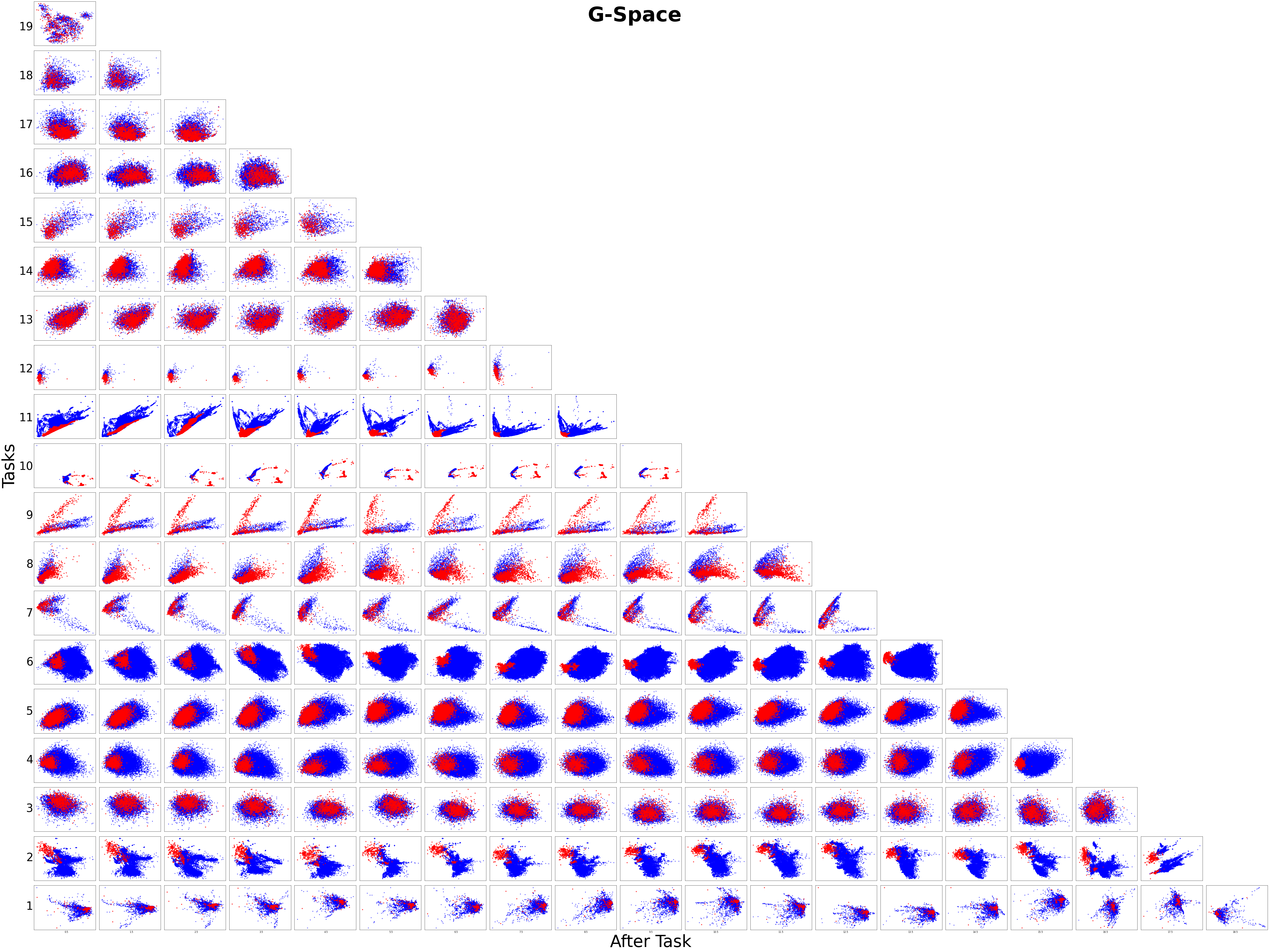}
\caption{Evolution of the learned latent $G$-space across the continual learning sequence. 
Each row corresponds to a task, while columns show the embedding of the same task after subsequent tasks are learned. 
Blue and red points denote inliers and outliers, respectively. 
The overall geometry of previously learned tasks remains stable over time, and inlier–outlier separation is preserved, indicating minimal representation drift and robustness against catastrophic forgetting.}
\label{fig:gspace_evolution}
\end{figure}

\autoref{fig:gspace_evolution} visualizes the evolution of the learned latent $G$-space across all tasks. Each row corresponds to a task learned at a particular stage, while columns show the representation of the same task after subsequent tasks have been introduced. Blue and red points denote inliers and outliers respectively.
Several consistent patterns can be observed. First, the global geometry of previously learned tasks remains largely preserved as new datasets are incorporated. Earlier task manifolds do not collapse or shift drastically, indicating that the encoder maintains stable feature representations throughout training. This behavior contrasts with typical continual fine-tuning scenarios, where representations drift significantly as new tasks overwrite earlier knowledge.
Second, inlier and outlier structures remain well separated across later evaluations. In many cases, the separation becomes progressively clearer after additional tasks are learned, suggesting positive transfer between datasets. Exposure to diverse anomaly patterns appears to refine the shared latent manifold rather than introducing interference.
Third, the absence of abrupt geometric changes across columns indicates minimal representation drift. Once a task representation is formed, its embedding remains visually consistent even after many subsequent updates, providing qualitative evidence that the proposed alignment and augmentation mechanisms effectively mitigate catastrophic forgetting.

\subsection{Anomaly Score Stability Across Tasks}

\begin{figure}[]
\centering
\includegraphics[width=\linewidth]{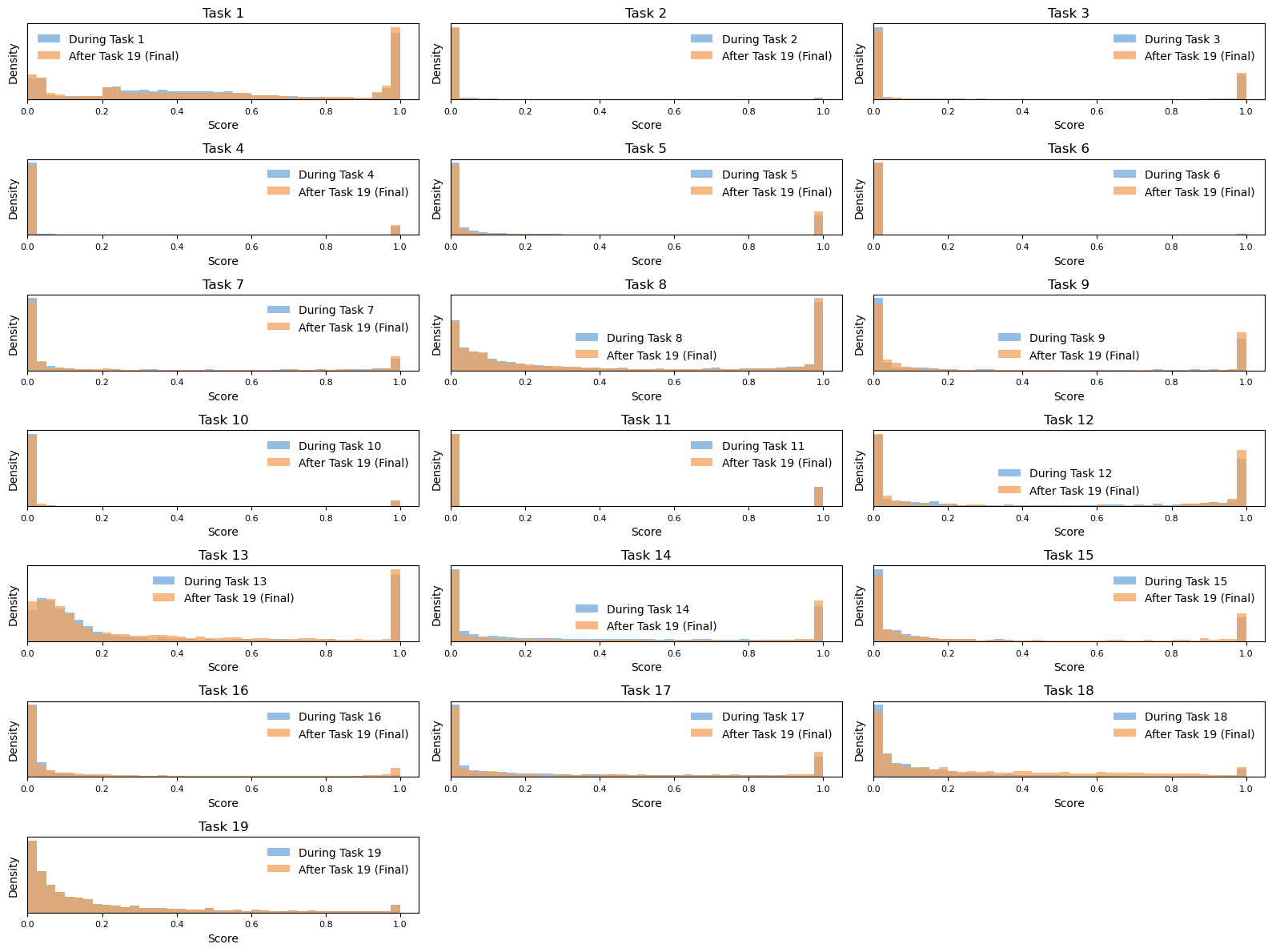}
\caption{Anomaly score stability across tasks. 
For each task, histograms compare anomaly score distributions at learning time (blue) and after completion of the full task sequence (orange). 
The strong overlap between distributions indicates that decision boundaries remain stable over continual updates, demonstrating preservation of previously learned anomaly detection behavior.}
\label{fig:score_stability}
\end{figure}

\autoref{fig:score_stability} analyzes the stability of anomaly scores by comparing score distributions at task learning time and after completion of the full task sequence. For each task, histograms show the distribution of anomaly scores computed immediately after learning (blue) and after the final task (orange).
Across all tasks, the two distributions exhibit strong overlap, indicating that the scoring function remains stable over time. Importantly, previously learned tasks do not experience distributional shifts toward higher or lower anomaly scores, suggesting that decision boundaries remain consistent despite continued training. This stability implies that newly learned tasks do not overwrite earlier anomaly detection behavior.
In several cases, score distributions become slightly sharper after later tasks, which suggests improved calibration rather than degradation. This observation supports the hypothesis that continual exposure to heterogeneous datasets helps refine shared anomaly representations, leading to incremental knowledge accumulation.

Together (\autoref{fig:gspace_evolution} and \autoref{fig:score_stability}), these analyses provide complementary qualitative and statistical evidence supporting the robustness of the proposed framework. The latent-space visualization demonstrates structural stability of learned representations, while the score-distribution analysis confirms functional stability of anomaly predictions. These results reinforce the main experimental findings by showing that continual learning proceeds through progressive refinement rather than destructive interference.

\end{document}